\pdfoutput=1
\documentclass[letterpaper, 10 pt, conference]{ieeeconf}  

\IEEEoverridecommandlockouts                              
\usepackage{graphicx}
\usepackage{amsmath} \usepackage{amssymb}  
\usepackage{booktabs}
\usepackage{multirow}
\usepackage{microtype}

\title{\LARGE \bf
TrajWAM: An Efficient Trajectory-Guided World Action Model for Dynamic Robotic Manipulation
}

\author{Haohan Geng$^{1,\dagger}$,
Sha Zhang$^{2,\dagger}$,
Weishuo Sun$^{1,\dagger}$,
Guanting Ye$^{3}$,
Wenhao Yu$^{1}$,\\[3pt]
Yuang Xie$^{1}$,
Qicheng Zeng$^{4}$,
Peng Ye$^{2}$,
Jiajun Deng$^{1}$,
Jianmin Ji$^{1,*}$\thanks{$^{1}$University of Science and Technology of China.}\thanks{$^{2}$The Chinese University of Hong Kong.}\thanks{$^{3}$University of Macau.}\thanks{$^{4}$Huawei Technologies Co., Ltd.}\thanks{$^{\dagger}$These authors contributed equally to this work.}\thanks{$^{*}$Corresponding author.}}

\begin{document}

\maketitle
\thispagestyle{empty}
\pagestyle{empty}

\begin{abstract}

Dynamic object manipulation requires robots to anticipate target motion while replanning fast enough to avoid executing outdated actions. 
Existing approaches either rely on fast reactive control without an explicit future target state or encode anticipated motion only implicitly.
World Action Models (WAMs) offer a promising alternative by coupling future visual prediction with action generation, but scene-level video objectives may underrepresent target dynamics that are critical for manipulation, while iterative video-action denoising introduces substantial latency. 
We present \textbf{TrajWAM}, a trajectory-guided world action model for efficient dynamic robotic manipulation. 
Our key insight is to explicitly ground future visual imagination in predicted target-object motion and use the resulting future representation to condition action generation, while preserving this anticipation pathway after acceleration.
Given a short history of object masks, a history-conditioned motion predictor forecasts future masks encoding the target object's trajectory. 
A lightweight Trajectory ControlNet then steers a pretrained video model toward trajectory-consistent futures. The resulting future representation conditions an action expert to predict executable action chunks.
To enable one-step inference without breaking this interface, we introduce a decoupled distillation scheme that combines distribution matching for the video backbone, Trajectory ControlNet realignment, and action rollout distillation to support accurate one-step action prediction conditioned on the compressed future representation.
Comprehensive evaluations in simulation and the real world show that TrajWAM achieves state-of-the-art performance on dynamic object manipulation tasks while enabling efficient one-step video-action inference.
Project page: https://kevingeng3.github.io/TrajWAM/.

\end{abstract}

\section{Introduction}
\label{sec:introduction}

Dynamic object manipulation is essential for robots operating in human environments, factories, and warehouses, where task-relevant objects may continue moving while the robot perceives and acts~\cite{10948321, c2rope, sope}. Unlike static manipulation, in which the target state is approximately fixed between observation and execution, dynamic manipulation couples perception, prediction, and control in a continuously changing scene. An action that is appropriate for the current observation may therefore become outdated before execution, and even a small delay may lead to a missed grasp or inaccurate contact. Successful dynamic manipulation consequently requires both anticipation of the target object's motion and sufficiently fast closed-loop replanning.

\begin{figure}[t]
    \setlength{\abovecaptionskip}{-0.2em}
    \centering
    \resizebox{\linewidth}{!}{\includegraphics{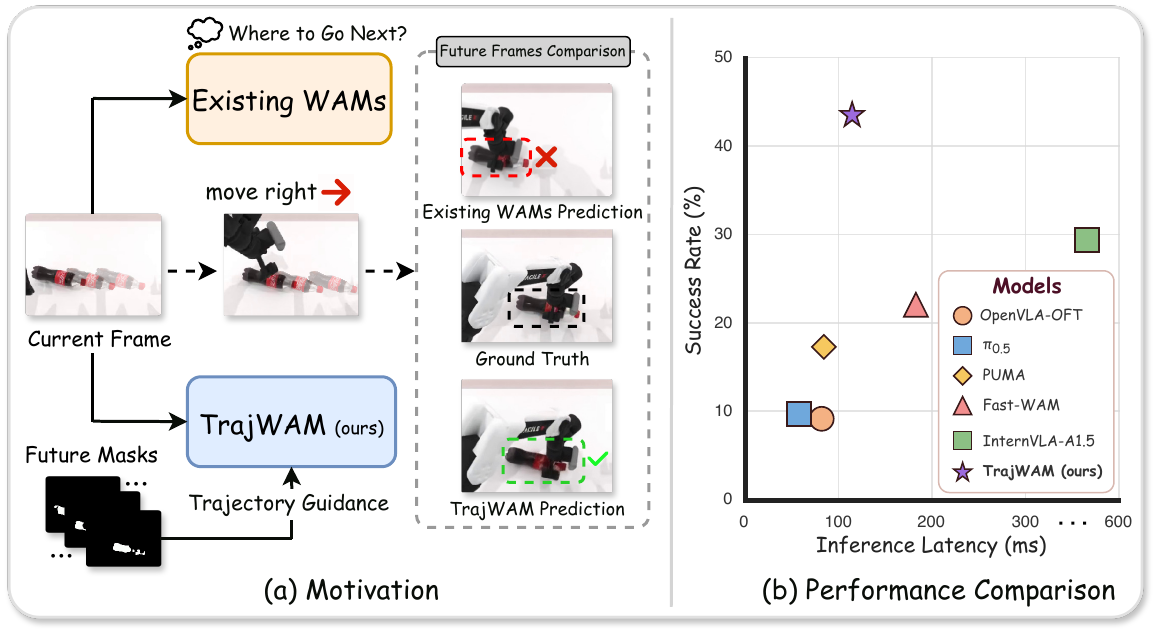}}
    \caption{Motivation and performance comparison. (a) Existing WAMs may fail to accurately anticipate target-object motion, resulting in future predictions that deviate from the actual trajectory. TrajWAM uses predicted future object masks as trajectory guidance to generate motion-consistent futures for action prediction. (b) Success rate and inference latency on DOMINO. TrajWAM achieves state-of-the-art performance while maintaining efficient inference.}
    \label{fig:teaser}
    \vspace{-1.0em}
\end{figure}

World Action Models (WAMs) have recently emerged as a promising alternative to vision-language-action policies by coupling future visual prediction with action generation ~\cite{bi2025motusunifiedlatentaction, ye2026worldactionmodelszeroshot, li2026causalworldmodelingrobot}. 
By modeling how observations evolve under interaction, WAMs provide an additional physical prior for control and are particularly appealing for dynamic manipulation.  
However, as illustrated in Fig.~\ref{fig:teaser}, 
existing WAMs predict scene evolution at the full-frame level, where robot-induced changes can dominate the imagined future.
In dynamic manipulation, task success may instead depend critically on target motion that evolves independently of the robot's actions. 
A scene-level forecast entangles robot motion, background dynamics, and target-object motion, making the task-relevant trajectory only indirectly represented and difficult to control explicitly during generation.
Extending WAMs to dynamic object manipulation thus calls for an object-centric future representation that specifies the anticipated spatial trajectory of the target while remaining coupled to action prediction.

Moreover, most existing WAMs synthesize future videos and actions through iterative denoising, introducing latency that can make the resulting prediction stale before execution. Recent methods reduce this cost either by skipping future imagination during inference~\cite{yuan2026fastwamworldactionmodels, ye2026gigaworldpolicyefficientactioncenteredworldaction} or by compressing the denoising process~\cite{li2026efficientwam1bparameterworldactionmodel, akbari2026flashwammodalityawaredistillationworld, cai2026ahawamasynchronoushorizonadaptiveworldactionmodeling}. However, the former sacrifices an explicit predictive pathway for anticipating target motion, while the latter leaves open whether a spatially explicit motion condition can remain effective after aggressive denoising compression.

To address these challenges, we propose \textbf{TrajWAM}, a trajectory-guided world action model for efficient dynamic robotic manipulation. 
Our key insight is to explicitly incorporate predicted target-object motion into future visual imagination and propagate its influence to action generation through the resulting video representation. 
Starting from a short observation history, TrajWAM predicts how the task-relevant object will evolve and represents this motion as a sequence of future object masks. These predicted masks provide an explicit spatial guide for future generation, steering the world model toward futures that are consistent with the anticipated target motion. 
The resulting imagined future then conditions action prediction, allowing the robot to act not only on what is currently observed, but also on where the target is expected to be.

The remaining challenge is to make this anticipation pathway fast enough for dynamic control. 
Simply bypassing future imagination would remove the mechanism through which predicted target motion influences the action, while naively compressing the generative process can degrade both trajectory controllability and downstream action prediction. 
We therefore introduce a decoupled one-step distillation strategy that accelerates the video and action generation processes according to their different requirements. 
The video pathway is distilled while explicitly preserving its response to trajectory guidance, whereas the action pathway is distilled to retain accurate predictions under the compressed future representation. 
In this way, TrajWAM substantially reduces inference latency without discarding the trajectory-guided future imagination that enables anticipatory control.

We evaluate TrajWAM on dynamic manipulation tasks in both simulated and real-world settings, covering diverse object categories, motion patterns, and task types. On DOMINO, as shown in Fig.~\ref{fig:teaser}(b), TrajWAM achieves state-of-the-art performance with a success rate of 43.54\%, outperforming the strongest evaluated baseline by 14.24 percentage points at an inference latency of 108.5\, ms. 
In real-world experiments, TrajWAM achieves success rates of 77.9\% on conveyor-belt grasping and 70.4\% on rotating-disc grasping, consistently outperforming the evaluated baselines.

Our contributions are summarized as follows:

\begin{itemize}
    \item We introduce TrajWAM, a trajectory-guided world action model that explicitly incorporates predicted target-object motion into future visual imagination and propagates this anticipation to downstream action prediction.
    \item We develop a decoupled one-step distillation strategy that accelerates video and action generation while preserving trajectory controllability and accurate future-conditioned action prediction.
    \item Extensive experiments in simulation and on real robots establish state-of-the-art dynamic manipulation performance while demonstrating efficient one-step inference and the effectiveness of trajectory-guided anticipation.
\end{itemize}

\section{Related Work}

\paragraph{World Action Models and Dynamic Robotic Manipulation}
Vision-language-action models have made substantial progress in robot manipulation by directly mapping visual observations and language instructions to actions~\cite{kim2024openvlaopensourcevisionlanguageactionmodel,black2026pi0visionlanguageactionflowmodel}. World-action models (WAMs) extend this paradigm by jointly modeling future observations and executable actions, using visual dynamics prediction as an additional physical prior for control. Recent WAMs have studied different ways to balance future imagination and action execution. Fast-WAM~\cite{yuan2026fastwamworldactionmodels} shows that explicit future generation can be skipped at inference, while AHA-WAM~\cite{cai2026ahawamasynchronoushorizonadaptiveworldactionmodeling} and HarmoWAM~\cite{feng2026harmowamharmonizinggeneralizableprecise} further improve the temporal coordination between world prediction and action generation. Dynamic manipulation requires anticipating the motion of objects that move independently of the robot. DOMINO~\cite{fang2026generalizableroboticmanipulationdynamic} exposes the resulting performance degradation, while DynamicVLA~\cite{xie2026dynamicvlavisionlanguageactionmodeldynamic} and PUMA~\cite{fang2026generalizableroboticmanipulationdynamic} address it through low-latency inference and implicit object-motion prediction, respectively. DynamicWAM~\cite{lou2026dynamicwamdualpathmotionconditioning} incorporates recent motion into WAMs through dual-path flow conditioning, whereas TrajWAM predicts future target-object mask trajectories to guide future imagination for action prediction.

\paragraph{Trajectory-Controllable Video Generation}
Trajectory-controllable video generation aims to synthesize videos that follow explicit spatial motion conditions, complementing text- or image-conditioned generation where object motion is often under-specified. Existing methods use point tracks or trajectories for fine-grained motion control~\cite{zhang2025toratrajectoryorienteddiffusiontransformer}, bounding boxes or tracklets to guide object locations~\cite{wang2024boximatorgeneratingrichcontrollable} and interactions~\cite{li2024trackdiffusiontrackletconditionedvideogeneration}, and 3D or depth-aware trajectories to reduce ambiguity in object motion~\cite{wang2025levitor3dtrajectoryoriented}. More recently, MagicMotion~\cite{li2025magicmotioncontrollablevideogeneration} proposes dense-to-sparse trajectory guidance with masks, boxes, and sparse boxes, improving trajectory adherence and object consistency. We build on this idea of explicit motion conditioning, but use it as a future-imagination mechanism inside WAMs: future object masks are predicted from observation history and injected into the video branch to support downstream dynamic manipulation.

\paragraph{Distillation for Diffusion-Based Models}
Dynamic manipulation requires low-latency inference, since the policy must continuously update actions while objects move. This is challenging for diffusion-based WAMs, where both future imagination and action generation rely on iterative denoising. Diffusion distillation provides a common way to reduce sampling steps~\cite{salimans2022progressivedistillationfastsampling,song2023consistencymodels,luo2023latentconsistencymodelssynthesizing}, and has been extended to efficient video generation~\cite{wang2023videolcmvideolatentconsistency,li2024t2vturbobreakingqualitybottleneck} and trajectory-controllable generation~\cite{li2026flashmotionfewstepcontrollablevideo}. Recent WAMs also explore this direction: Flash-WAM~\cite{akbari2026flashwammodalityawaredistillationworld} proposes modality-aware step distillation for joint video-action diffusion, while AHA-WAM-Flash~\cite{cai2026ahawamasynchronoushorizonadaptiveworldactionmodeling} applies ODE distillation to compress its action sampler from multi-step denoising to few-step inference. While these works primarily improve the efficiency of existing WAM inference pipelines, preserving trajectory controllability and action accuracy under compression remains underexplored.

\begin{figure*}[t]
    \setlength{\abovecaptionskip}{-0.2em}
    \centering
    \resizebox{\textwidth}{!}{\includegraphics{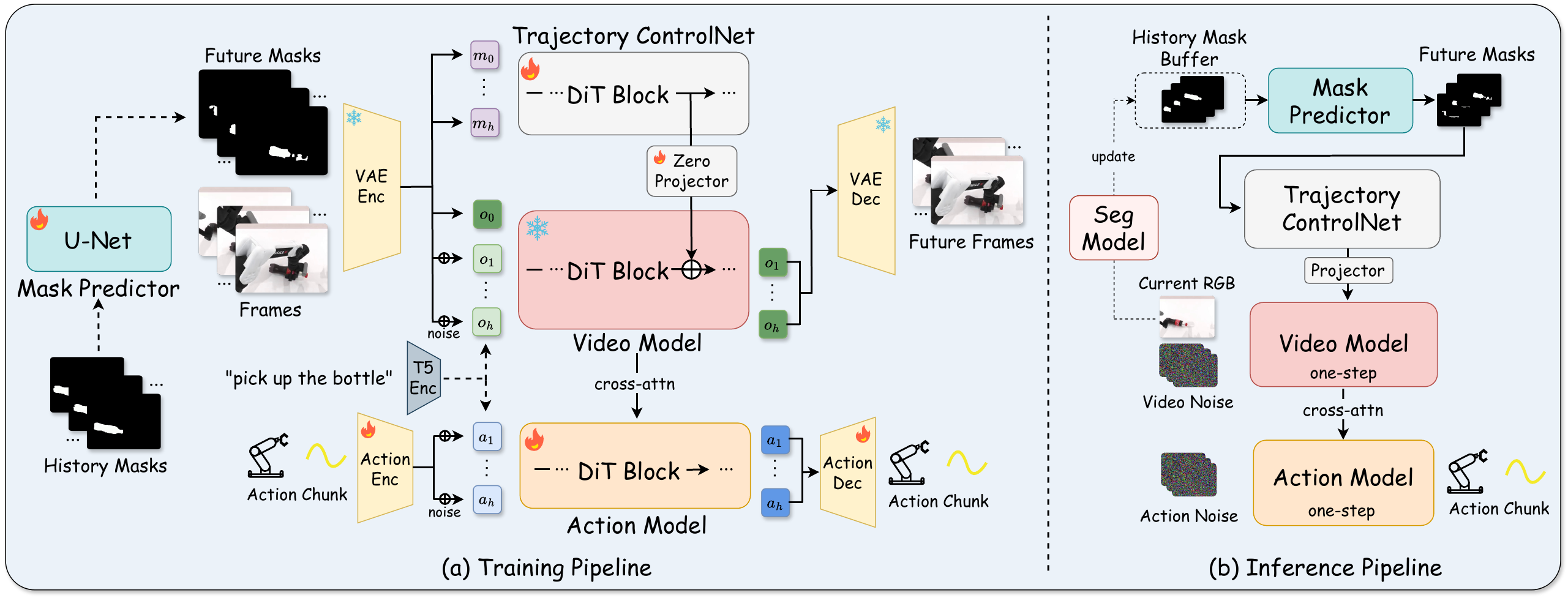}}
    \caption{Overview of TrajWAM. 
    (a) \textbf{Training pipeline.} A language-conditioned segmentation model extracts the mask history of the target object, from which a motion predictor forecasts a sequence of future masks representing its anticipated trajectory. The predicted mask trajectory guides future-video generation through a Trajectory ControlNet, and the resulting trajectory-controlled video representation conditions the action branch through asymmetric video--action attention to predict an executable action chunk.
    (b) \textbf{Inference pipeline.} Target-object masks are updated online from incoming observations and maintained in a short history buffer. The predicted future mask trajectory guides the distilled one-step video model, whose future representation conditions the one-step action model for low-latency closed-loop control.
    }
    \label{fig:model}
    \vspace{-1.0em}
\end{figure*}

\section{Method}
\label{sec:method}

TrajWAM connects anticipated target-object motion to action generation through controlled video generation. As shown in Fig.~\ref{fig:model}, given a language instruction and a short observation history, TrajWAM first obtains the mask history of the referred object and forecasts a sequence of future masks representing its anticipated trajectory. The predicted mask trajectory then guides future video generation, and the resulting trajectory-controlled video representation conditions the action expert to predict executable robot actions. However, the resulting world-action model remains too slow for dynamic manipulation, where excessive inference latency can make the predicted target state and corresponding action outdated before execution. We therefore distill TrajWAM into a one-step model to enable sufficiently fast closed-loop control for manipulating moving objects.

\subsection{Object-Centric Future Trajectory Prediction}
\label{sec:trajectory_prediction}

To provide the video model with object-centric dynamics, we use a frozen pretrained language-conditioned segmentation model $\mathcal S$ to extract the target-object mask $m_i=\mathcal S(o_i^{3rd},\ell)$ from the third-person camera's view, where $\ell$ denotes the language instruction. We use the fixed third-person view because it provides a stable global reference for observing the target-object trajectory. In contrast, the wrist-mounted cameras move with the robot arms and may intermittently lose sight of the target, making them unsuitable for maintaining a consistent mask history. The wrist views therefore provide RGB observations only and do not provide mask inputs.

Given a causal history $M_t^-$, which consists of $K$ masks sampled with temporal stride $s$ up to the current time $t$, a compact 2D U-Net~\cite{ronneberger2015unetconvolutionalnetworksbiomedical} $g_\phi$ predicts all $H_v$ future masks in one forward pass:
\begin{equation}
    M_t^+ = \sigma\!\left(g_\phi(M_t^-)\right),
\end{equation}
where $\sigma$ denotes the element-wise sigmoid and $M_t^+$ contains foreground probabilities. The predictor is trained with binary cross-entropy on the logits and soft Dice loss~\cite{milletari2016vnetfullyconvolutionalneural} on the predicted probabilities, each weighted by $0.5$, to balance pixel-wise foreground classification with region-level mask overlap.

\subsection{Trajectory-Conditioned Video--Action Modeling}
\label{sec:trajectory_wam}

The predicted mask trajectory provides explicit information about how the referred object will move. TrajWAM injects this trajectory into the video branch and lets the action branch read the resulting controlled video tokens through asymmetric joint self-attention.

\paragraph{Trajectory ControlNet}
As shown in Fig.~\ref{fig:model}(a), the predicted mask trajectory is encoded by the pretrained video VAE into trajectory latents $z_M=\mathcal E_{\mathrm{vae}}(M_t^+)$.
A lightweight Trajectory ControlNet~\cite{li2025magicmotioncontrollablevideogeneration} maps these latents and the video flow time to layer-wise residuals. Specifically, our video model is initialized from Wan2.2-5B~\cite{wan2025wanopenadvancedlargescale}, and the Trajectory ControlNet is randomly initialized while retaining the same depth as Wan2.2-5B but reducing its hidden width from 3,072 to 512. 
Conditioned on $z_M$ and the video flow time $s_v$, the Trajectory ControlNet injects residuals into each video block:
\begin{equation}
    \tilde h_v^{j}
    = h_v^{j}
    + \mathcal Z_j\!\left(A_\varphi^{j}(z_M,s_v)\right),
\end{equation}
where $h_v^j$ is the hidden state of the $j$-th video block, $A_\varphi^j$ is the corresponding ControlNet block, and $\mathcal Z_j$ is a zero-initialized convolutional projection.

\paragraph{Asymmetric Video--Action Attention}
The action expert is initialized from the pretrained Wan2.2-5B, retaining the same depth while reducing its hidden width from 3,072 to 1,024. At each transformer block, video and action tokens are concatenated for self-attention:
  \begin{equation}
    [h_v;h_a]
    =
    \operatorname{softmax}
    \!\left(
      \mathcal M(\frac{QK^\top}{\sqrt{d}})
    \right)
    V,
  \end{equation}
where an asymmetric attention mask \(\mathcal M(\cdot)\) controls the information flow: video queries attend only to video keys and values, whereas action queries attend to both action and video keys and values. Consequently, the action branch reads the trajectory-controlled future-video representation, while the video prediction remains independent of action tokens.

\paragraph{Training Objective}
During training, the video backbone, VAE, and text encoder are frozen, while the Trajectory ControlNet, zero-initialized projection layers, and action expert are updated.
Let $x_v=y_{>t}$ and $x_a=a_{>t}$ denote the clean future video latents and ground-truth action chunk, respectively. For each modality $q\in\{v,a\}$, we independently sample $\epsilon_q\sim\mathcal N(0,I)$ and $s_q\sim p(s)$, and construct noisy inputs $x_q^{s_q}=(1-s_q)x_q+s_q\epsilon_q$. Given these inputs, their flow times, and the condition $C=(y_t,z_M,c)$, the model predicts video and action velocities $\hat u_v$ and $\hat u_a$ with the objective
\begin{equation}
    \mathcal L
    = \mathbb E\!\left[
        \sum_{q\in\{v,a\}}
        \lambda_q
        \left\|\hat u_q-(\epsilon_q-x_q)\right\|_2^2
    \right],
\end{equation}
where $y_t$ and $c$ denote the current-frame latent and language condition, and $\lambda_v,\lambda_a$ balance action learning and video co-training. Thus, ground-truth future actions provide supervision for the action branch, while the trajectory condition influences action prediction through the controlled video prediction in the same joint forward pass.

\subsection{One-Step Deployment via Decoupled Distillation}
\label{sec:decoupled_one_step_distillation}
The trajectory-conditioned model above relies on iterative flow sampling for future video and action generation; its inference latency is difficult to reconcile with the control frequency required for closed-loop dynamic manipulation. Compressing both generation processes to a single step introduces two coupled changes: the video generator loses the refinement of multi-step sampling, and the action branch must read a different video representation. We therefore distill the model in three decoupled stages: first compress the uncontrolled video generator, then realign the Trajectory ControlNet with this one-step generator, and finally adapt the action branch to the deployed trajectory-conditioned generation process.

\begin{figure}[t]
    \setlength{\abovecaptionskip}{-0.2em}
    \centering
    \resizebox{\linewidth}{!}{\includegraphics{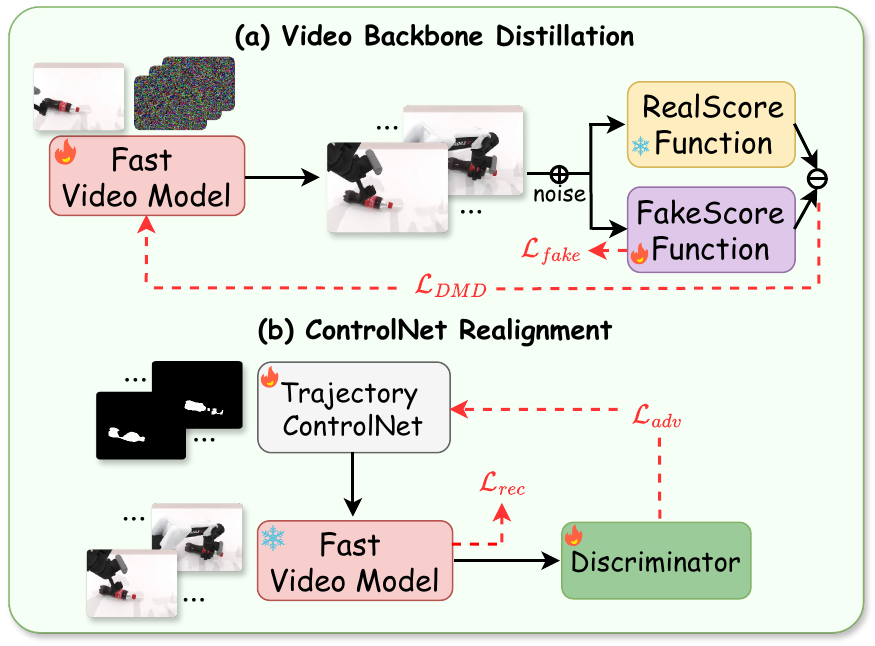}}
    \caption{Video distillation. (a) Video backbone distillation compresses the multi-step video generator into a fast generator using a frozen real-score function and a trainable fake-score function under distribution matching distillation. (b) ControlNet realignment adapts the Trajectory ControlNet to the frozen fast video model using latent reconstruction and diffusion-discriminator objectives.}
    \label{fig:video_distillation}
    \vspace{-1.0em}
\end{figure}

\paragraph{Video Backbone Distillation}
We first accelerate the video model independently of the Trajectory ControlNet and action branch, as illustrated in Fig.~\ref{fig:video_distillation}(a). The one-step generator $G_\psi$ produces future latents $\hat y^0=G_\psi(\epsilon_v\mid y_t,c)$ from Gaussian noise $\epsilon_v$, where $y_t$ denotes the current frame latent and $c$ denotes the language condition. 
Following distribution matching distillation (DMD)~\cite{yin2024onestepdiffusiondistributionmatching}, we perturb student samples as $\hat y^\tau=(1-\tau)\hat y^0+\tau\epsilon$, with a random noise level $\tau$ and $\epsilon\sim\mathcal N(0,I)$.
A frozen real-score model $f_\Theta$ and a trainable fake-score model $f_\omega$ yield clean predictions:
\begin{equation}
    \begin{aligned}
        \hat y_{\mathrm{real}}^0
        &= \hat y^\tau
        - \tau f_\Theta(\hat y^\tau,\tau\mid C_{\mathrm{vid}}),\\
        \hat y_{\mathrm{fake}}^0
        &= \hat y^\tau
        - \tau f_\omega(\hat y^\tau,\tau\mid C_{\mathrm{vid}}).
    \end{aligned}
\end{equation}
Their normalized difference defines the DMD direction
\begin{equation}
    \Delta_{\mathrm{DMD}}
    =
    \frac{
        \hat y_{\mathrm{fake}}^0-\hat y_{\mathrm{real}}^0
    }{
        \operatorname{mean}
        (|\hat y^0-\hat y_{\mathrm{real}}^0|)
        +\varepsilon
    },
\end{equation}
where $\varepsilon>0$ is a small positive constant added for numerical stability.
The generator is optimized with the surrogate objective
\begin{equation}
    \mathcal L_{\mathrm{DMD}}
    =
    \mathbb E\!\left[
        \frac12
        \left\|
            \hat y^0-
            \operatorname{sg}
            [\hat y^0-\Delta_{\mathrm{DMD}}]
        \right\|_2^2
    \right],
\end{equation}
where $\operatorname{sg}$ denotes stop-gradient.
The fake-score model minimizes the mean-squared error between $\hat y_{\mathrm{fake}}^0$ and the detached student sample $\operatorname{sg}[\hat y^0]$. We alternate generator and fake-score updates while keeping the real-score model frozen.

\paragraph{Trajectory Adapter Realignment}
The slow Trajectory ControlNet was trained with the intermediate feature
distribution of the multi-step video generator. After video backbone distillation, directly attaching it to the one-step generator causes a mismatch in both noise levels and intermediate activations. As illustrated in Fig.~\ref{fig:video_distillation}(b), we therefore initialize a fast trajectory adapter $\varphi_{\mathrm{fast}}$ from the trained slow Trajectory ControlNet $\varphi_{\mathrm{slow}}$ and realign it while keeping the fast video generator frozen. For a latent $y^\tau$ on the one-step schedule, the controlled generator predicts
\begin{equation}
  \hat y_{\mathrm{fast}}^{0}
  =
  y^\tau-\tau
  u_{\psi,\varphi_{\mathrm{fast}}}
  (y^\tau,\tau\mid y_t,z_M,c).
\end{equation}

We train the adapter with latent reconstruction and adversarial objectives. The diffusion discriminator $D_\eta$ uses a frozen pretrained multi-step video model with lightweight trainable attention modules at selected intermediate layers and a shallow MLP head. Conditioned on $z_M$ and $c$, it assigns logits $d_{\mathrm r}$ and $d_{\mathrm f}$ to ground-truth and generated future latents, respectively, after perturbing both with the same Gaussian noise and noise level.
The training objectives are
\begin{equation}
    \begin{aligned}
        \mathcal L_{\mathrm{rec}}
        &=
        \left\|
            \hat y_{\mathrm{fast}}^0-y_{>t}
        \right\|_2^2,\\
        \mathcal L_{\mathrm{adv}}
        &=
        \mathbb E\!\left[
            \operatorname{softplus}(-d_{\mathrm f})
        \right],\\
        \mathcal L_D
        &=
        \mathbb E\!\left[
            \operatorname{softplus}(-d_{\mathrm r})
            +\operatorname{softplus}(d_{\mathrm f})
        \right],
    \end{aligned}
\end{equation}
where $y_{>t}$ is the ground-truth future video latent. The adapter is updated using $\mathcal L_{\mathrm{rec}}$ and $\mathcal L_{\mathrm{adv}}$, while the discriminator is updated using $\mathcal L_D$.

\begin{figure}[t]
    \setlength{\abovecaptionskip}{-0.2em}
    \centering
    \resizebox{\linewidth}{!}{\includegraphics{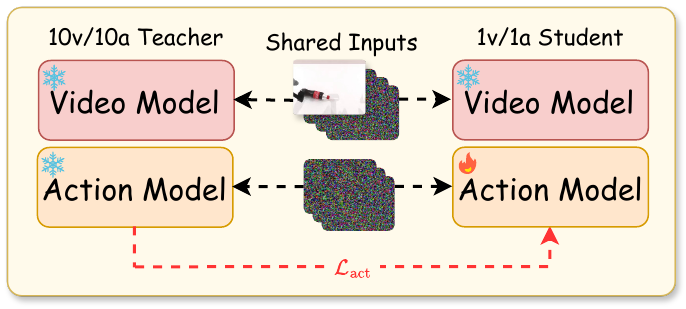}}
    \caption{Action rollout distillation. The slow teacher and fast student receive the same observation, language instruction, trajectory condition, and initial noises. The trainable fast action model is optimized to match the action chunk generated by the slow multi-step teacher.}
    \label{fig:action_distillation}
    \vspace{-1.0em}
\end{figure}

\paragraph{Action Rollout Distillation}
After the one-step video generator and trajectory adapter are fixed, we adapt the action branch to the deployed one-step trajectory-conditioned student, as illustrated in Fig.~\ref{fig:action_distillation}. Let $\Phi_{\theta_T}^{10,10}$ and $\Phi_{\theta_S}^{1,1}$ denote the teacher and student video-action rollouts, where the superscripts indicate the numbers of video and action denoising steps. They receive the same observation, language instruction, trajectory condition, and initial video/action noises:
\begin{align}
  (\hat y_T,\hat a_T)
  &=
  \Phi_{\theta_T}^{10,10}
  (\epsilon_v,\epsilon_a\mid y_t,z_M,c),\\
  (\hat y_S,\hat a_S)
  &=
  \Phi_{\theta_S}^{1,1}
  (\epsilon_v,\epsilon_a\mid y_t,z_M,c).
\end{align}
The student is trained to match the action chunk produced by the teacher's
multi-step rollout, rather than an intermediate flow velocity or a ground-truth
action-flow target. For a minibatch of $B$
samples and an action chunk of $T_a$ tokens with dimension $D_a$, define
\begin{equation}
  \ell_{b,t}
  =
  \frac{1}{D_a}
  \sum_{d=1}^{D_a}
  \left(
    \hat a_{S,b,t,d}
    -
    \operatorname{sg}(\hat a_{T,b,t,d})
  \right)^2,
\end{equation}
where $\operatorname{sg}(\cdot)$ stops gradients through the teacher. The training objective is
\begin{equation}
  \mathcal L_{\mathrm{act}}
  =
  \frac{1}{B}
  \sum_{b=1}^{B}
  \frac{
    \sum_{t=1}^{T_a}m_{b,t}\ell_{b,t}
  }{
    \max(1,\sum_{t=1}^{T_a}m_{b,t})
  },
\end{equation}
where $m_{b,t}$ masks padded action positions.
\section{Experiments}

\subsection{Experimental Setup}
\label{Experimental Setup}
\subsubsection{Simulation Benchmark}
We evaluate TrajWAM on DOMINO, a dynamic manipulation benchmark, which contains 35 tasks spanning different object categories, motion patterns, and manipulation objectives~\cite{fang2026generalizableroboticmanipulationdynamic}. All methods are evaluated in a clean setting with the Aloha-AgileX robot, using a training mixture of 35 dynamic tasks under Level 1 dynamics with a dynamic coefficient of $\alpha = 0.1$. The training dataset contains 50 demonstrations per task, for a total of 1,750 demonstrations.

\subsubsection{Real-World Setup}
\begin{figure}[t]
    \setlength{\abovecaptionskip}{-0.2em}
    \centering
    \resizebox{\linewidth}{!}{\includegraphics{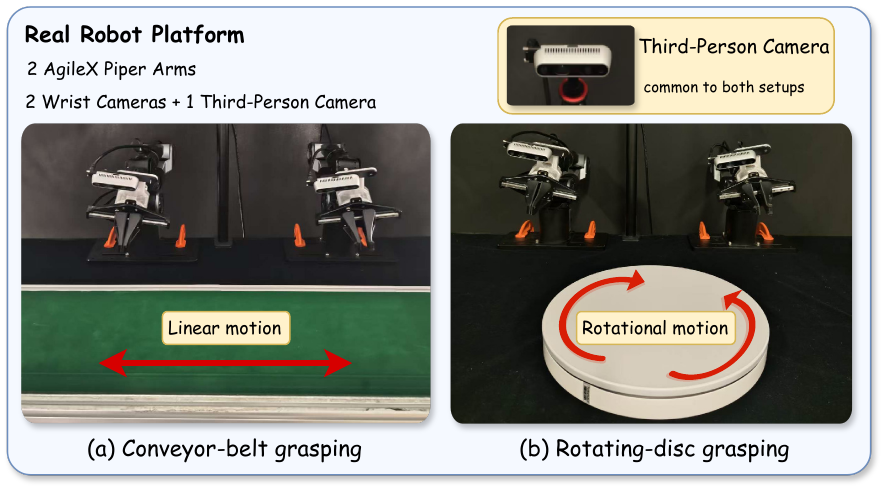}}
    \caption{Real-world evaluation setups. We evaluate TrajWAM on (a) conveyor-belt grasping with linear motion and (b) rotating-disc grasping with rotational motion. Both setups use two 6-DoF AgileX Piper arms, each equipped with a wrist-mounted Intel RealSense D435i camera, and a static D435i camera providing a global third-person view.}
    \label{fig:realworld}
    \vspace{-1.0em}
\end{figure}

We further evaluate TrajWAM on two real-world dynamic grasping tasks: conveyor-belt grasping and rotating-disc grasping, as illustrated in Fig.~\ref{fig:realworld}. Both tasks require anticipating the target's future motion before the grasp is executed. The platform consists of two 6-DoF AgileX Piper arms, each equipped with a wrist-mounted Intel RealSense D435i RGB camera, and a static D435i that provides the global third-person view. Each task is evaluated at three motion speeds (low, medium, and high) using objects from four categories (banana, bottle, tape, and plush toy), covering rolling and sliding contact behaviors. We collect 200 demonstrations per task (400 in total) through teleoperation. Each demonstration consists of a single successful grasp at a randomly sampled speed using a randomly selected object. For evaluation, we run 20 trials per (task, speed, object) combination with randomized initial poses, and count a trial as successful only if the object is grasped and lifted.

\begin{figure*}[t]
    \setlength{\abovecaptionskip}{-0.2em}
    \centering
    \resizebox{\textwidth}{!}{\includegraphics{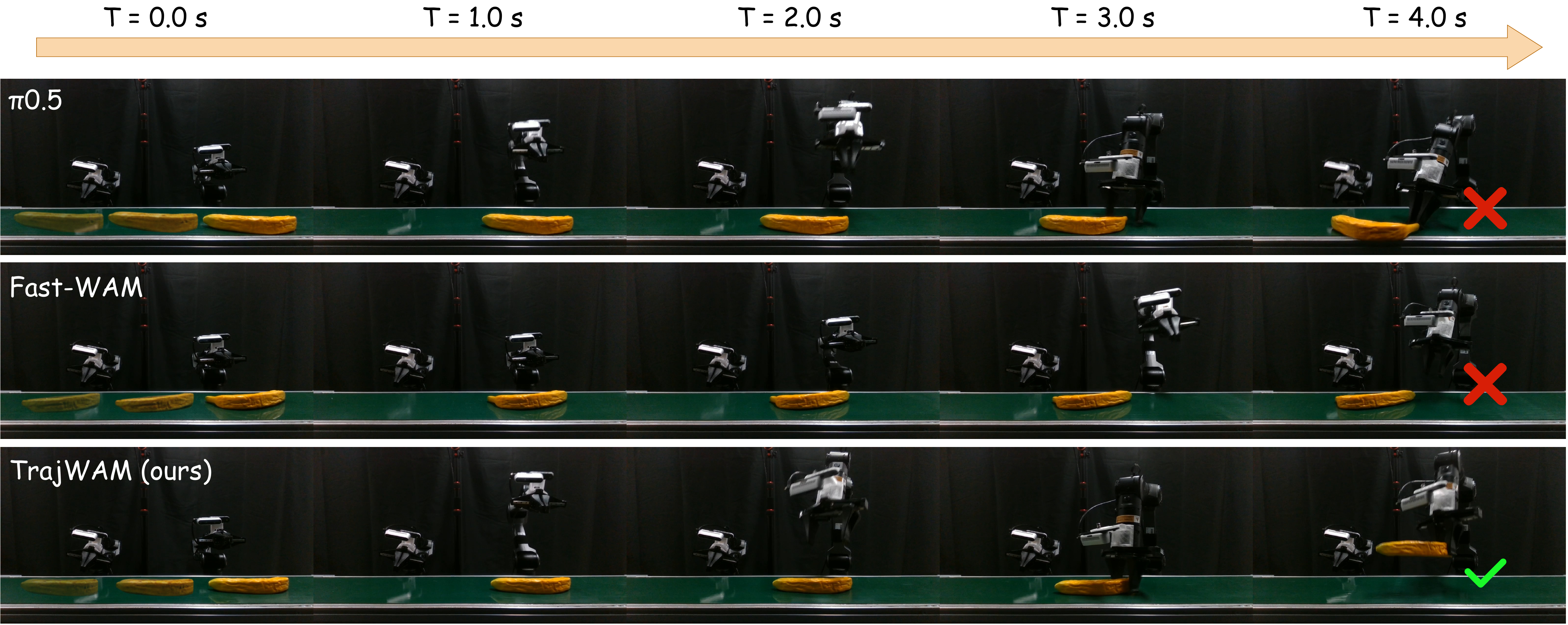}}
    \caption{Qualitative comparison of real-world conveyor-belt grasping. Snapshots are shown at 1\,s intervals from $T=0$ to $4$\,s. TrajWAM conditions action generation on trajectory-guided future visual representations to anticipate target motion, successfully grasping and lifting the moving banana. In the illustrated trials, $\pi_{0.5}$ and Fast-WAM fail to secure the object. Red crosses and the green check mark indicate grasp failure and success, respectively.}
    \label{fig:real_comp}
    \vspace{-1.0em}
\end{figure*}

\subsubsection{Implementation Details}
At inference, target-object masks are obtained using frozen SAM3~\cite{carion2026sam3segmentconcepts} and updated online during closed-loop control. In our real-world deployment, SAM3 runs asynchronously with a mean latency of 69.1 ms. At each replanning cycle, the policy uses the latest available mask history without waiting for segmentation of the current observation, while buffered actions continue to execute. The trajectory predictor takes $K=4$ historical masks at temporal stride $s=4$ and predicts $H_v=8$ future masks. It is trained for 20 epochs with the BCE--Dice loss (batch size 128, initial learning rate $10^{-3}$).
We initialize the video model and action model from pretrained Fast-WAM-Joint~\cite{yuan2026fastwamworldactionmodels}, which uses a video backbone derived from Wan2.2-5B. During trajectory-conditioned training, we freeze the video backbone, VAE, and text encoder while optimizing the Trajectory ControlNet and action expert for 5,000 steps using a global batch size of 128 and a learning rate of $5\times10^{-5}$.
Following Sec.~\ref{sec:decoupled_one_step_distillation}, we distill the video generator for 5,000 steps (global batch size 8), with generator/fake-score learning rates of $5\times10^{-7}$/$1\times10^{-7}$ and a 1:5 generator-to-fake-score update ratio. With the distilled video generator frozen, we realign the trajectory adapter for 1,000 steps (learning rate $2\times10^{-6}$), then distill the action branch for 2,000 steps (constant learning rate $10^{-6}$), both with a global batch size of 4. The teacher uses 10 denoising steps per modality; the deployed student uses one.
We use NVIDIA H200 GPUs for training, A800 GPUs for DOMINO evaluation, and a single NVIDIA RTX 5090 32 GB GPU for real-world experiments and all latency measurements. In both evaluation settings, the policy predicts $T_a=32$ actions and executes 8 before obtaining a new observation and replanning. All real-world policies use Real-Time Chunking (RTC)~\cite{black2025realtimeexecutionactionchunking}, while DOMINO evaluation retains the synchronous protocol.

\subsubsection{Evaluation Metrics}
For simulation experiments, we report success rate (SR), manipulation score(MS)~\cite{fang2026generalizableroboticmanipulationdynamic}, and inference latency. SR measures the percentage of successful task executions. MS is a continuous metric defined by DOMINO to capture manipulation execution quality, providing a finer-grained evaluation than binary task success. For real-world experiments, we report task success rate only.

\subsection{Main Results}
\subsubsection{Simulation Results}

  \begin{table}[t]
  \setlength{\belowcaptionskip}{-0.2em}
  \caption{Comparison with SOTA methods on DOMINO.}
  \label{table_domino}
      \centering
      \begin{tabular}{lccc}
          \hline
          \textbf{Method} & \textbf{SR (\%)} $\uparrow$ & \textbf{MS}  & \textbf{Latency (ms)} \\
          \hline
          OpenVLA~\cite{kim2024openvlaopensourcevisionlanguageactionmodel} & 1.54 & 6.10 & 173.6\\

          $\pi_0$-FAST~\cite{pertsch2025fastefficientactiontokenization} & 3.54 & 20.87 & 119.9\\

          VLA-Adapter~\cite{wang2025vlaadaptereffectiveparadigmtinyscale} & 4.40 & 24.31 & 83.6\\

          RDT-1B~\cite{liu2025rdt1bdiffusionfoundationmodel} & 5.34 & 17.71 & 246.3\\

          InternVLA-M1~\cite{chen2025internvlam1spatiallyguidedvisionlanguageaction} & 5.40 & 27.57 & 183.8\\

          Isaac-GR00T~\cite{nvidia2025gr00tn1openfoundation} & 6.10 & 28.60 & 76.4\\

          $\pi_0$~\cite{black2026pi0visionlanguageactionflowmodel} & 8.17 & 23.96 & 52.9\\

          OpenVLA-OFT~\cite{kim2025finetuningvisionlanguageactionmodelsoptimizing} & 9.06 & 24.06 & 83.1\\

          $\pi_{0.5}$~\cite{intelligence2025pi05visionlanguageactionmodelopenworld} & 9.63 & 26.17 & 59.1\\

          PUMA~\cite{fang2026generalizableroboticmanipulationdynamic} & 17.20 & 34.97 & 85.4\\

          Fast-WAM~\cite{yuan2026fastwamworldactionmodels} & 21.97 & 37.27 &183.7\\

          InternVLA-A1.5~\cite{ma2026internvlaa15unifyingunderstandinglatent} & \underline{29.30} & \underline{42.50} & 552.2\\
          
          TrajWAM (ours) & \textbf{43.54} & \textbf{54.07} & 108.5\\
          \hline
      \end{tabular}
      \vspace{-1.0em}
  \end{table}

Table~\ref{table_domino} compares TrajWAM with representative VLA and dynamic manipulation methods on DOMINO. TrajWAM achieves the best overall performance, with a success rate of 43.54\% and a manipulation score of 54.07. Compared with the strongest baseline, InternVLA-A1.5, TrajWAM improves SR by 14.24 percentage points and MS by 11.57 points, demonstrating the effectiveness of explicitly modeling future target-object motion.

TrajWAM achieves these gains with an inference latency of 108.5 ms. Despite introducing explicit future trajectory prediction and controlled video generation, it is faster than Fast-WAM (183.7 ms) and substantially faster than InternVLA-A1.5 (552.2 ms), while achieving considerably higher manipulation performance. Although several VLA baselines have lower latency, they exhibit substantially lower success rates on dynamic tasks. These results indicate that TrajWAM provides a favorable trade-off between dynamic manipulation capability and inference efficiency.

\subsubsection{Real-World Results}

\begin{table*}[t]
\setlength{\belowcaptionskip}{-0.2em}
\centering
\caption{Real-world task success rates.}
\label{tab:real_world_results}

\setlength{\tabcolsep}{2.5pt}

\begin{tabular*}{\textwidth}{
    @{\extracolsep{\fill}}l*{14}{c}@{}
}
\toprule

\multirow{2}{*}{Method}
& \multicolumn{4}{c}{Low-Speed}
& \multicolumn{4}{c}{Medium-Speed}
& \multicolumn{4}{c}{High-Speed}
& \multicolumn{2}{c}{Overall} \\

\cmidrule(lr){2-5}
\cmidrule(lr){6-9}
\cmidrule(lr){10-13}
\cmidrule(lr){14-15}

& Banana & Bottle & Tape & Plush toy
& Banana & Bottle & Tape & Plush toy
& Banana & Bottle & Tape & Plush toy
& Succ. & SR (\%) \\

\midrule

\multicolumn{15}{l}{\textit{Conveyor-belt}} \\
\addlinespace[2pt]

$\pi_{0.5}$
& \underline{15/20} & \textbf{18/20} & \underline{13/20} & \underline{19/20}
& \underline{9/20}  & \underline{10/20} & \underline{8/20}  & \underline{16/20}
& \underline{3/20}  & 0/20  & \underline{5/20}  & \underline{14/20}
& \underline{130/240} & \underline{54.2} \\

Fast-WAM
& 14/20 & 6/20  & 11/20 & 18/20
& 5/20  & 2/20  & 3/20  & 8/20
& 1/20  & 0/20  & 1/20  & 2/20
& 71/240 & 29.6 \\

TrajWAM (ours)
& \textbf{18/20} & \underline{16/20} & \textbf{15/20} & \textbf{20/20}
& \textbf{16/20} & \textbf{13/20} & \textbf{18/20} & \textbf{19/20}
& \textbf{11/20} & \textbf{7/20}  & \textbf{16/20} & \textbf{18/20}
& \textbf{187/240} & \textbf{77.9} \\

\midrule

\multicolumn{15}{l}{\textit{Rotating-disc}} \\
\addlinespace[2pt]

$\pi_{0.5}$
& \underline{12/20} & \underline{5/20} & \underline{8/20} & \underline{17/20}
& \underline{6/20} & \underline{4/20} & \underline{11/20} & \underline{15/20}
& \underline{5/20} & 0/20 & \underline{10/20} & \underline{13/20}
& \underline{106/240} & \underline{44.2} \\

Fast-WAM
& 4/20 & 1/20 & 6/20 & 16/20
& 2/20 & 2/20 & 9/20 & 14/20
& 1/20 & 0/20 & 2/20 & 10/20
& 67/240 & 27.9 \\

TrajWAM (ours)
& \textbf{14/20} & \textbf{11/20} & \textbf{16/20} & \textbf{18/20}
& \textbf{12/20} & \textbf{13/20} & \textbf{12/20} & \textbf{17/20}
& \textbf{12/20} & \textbf{10/20} & \textbf{15/20} & \textbf{19/20}
& \textbf{169/240} & \textbf{70.4} \\

\bottomrule
\end{tabular*}

\vspace{-1.0em}
\end{table*}

Table~\ref{tab:real_world_results} compares TrajWAM with $\pi_{0.5}$ and Fast-WAM on two real-world dynamic grasping tasks. TrajWAM achieves success rates of 77.9\% on conveyor-belt grasping and 70.4\% on rotating-disc grasping, outperforming $\pi_{0.5}$ (54.2\% and 44.2\%) and Fast-WAM (29.6\% and 27.9\%). Aggregating all 480 trials per method across both tasks yields an overall success rate of 74.2\% for TrajWAM, compared with 49.2\% for $\pi_{0.5}$ and 28.8\% for Fast-WAM. When results are pooled across the four object categories, TrajWAM also achieves the highest success rate in all six task–speed conditions.

The advantage remains pronounced under high-speed motion, where greater target displacement during action execution makes grasping more sensitive to timing and spatial alignment. TrajWAM achieves high-speed success rates of 65.0\% and 70.0\% on conveyor-belt and rotating-disc grasping, respectively, exceeding the strongest baseline, $\pi_{0.5}$, by 37.5 and 35.0 percentage points. These results are consistent with the benefit of conditioning action generation on trajectory-guided future representations, which provide anticipated target locations for grasp planning. Performance nevertheless varies across objects: high-speed conveyor-belt grasping remains challenging for bottles, with TrajWAM succeeding in 7 of 20 trials.

The real-world results further complement the DOMINO evaluation. Although Fast-WAM outperforms $\pi_{0.5}$ in simulation, their ranking reverses on both physical tasks, indicating that baseline rankings do not necessarily carry over across evaluation settings. TrajWAM maintains its advantage in both settings, supporting its effectiveness across the evaluated dynamic manipulation tasks.

\begin{figure}[t]
    \setlength{\abovecaptionskip}{-0.2em}
    \centering
    \resizebox{\linewidth}{!}{\includegraphics{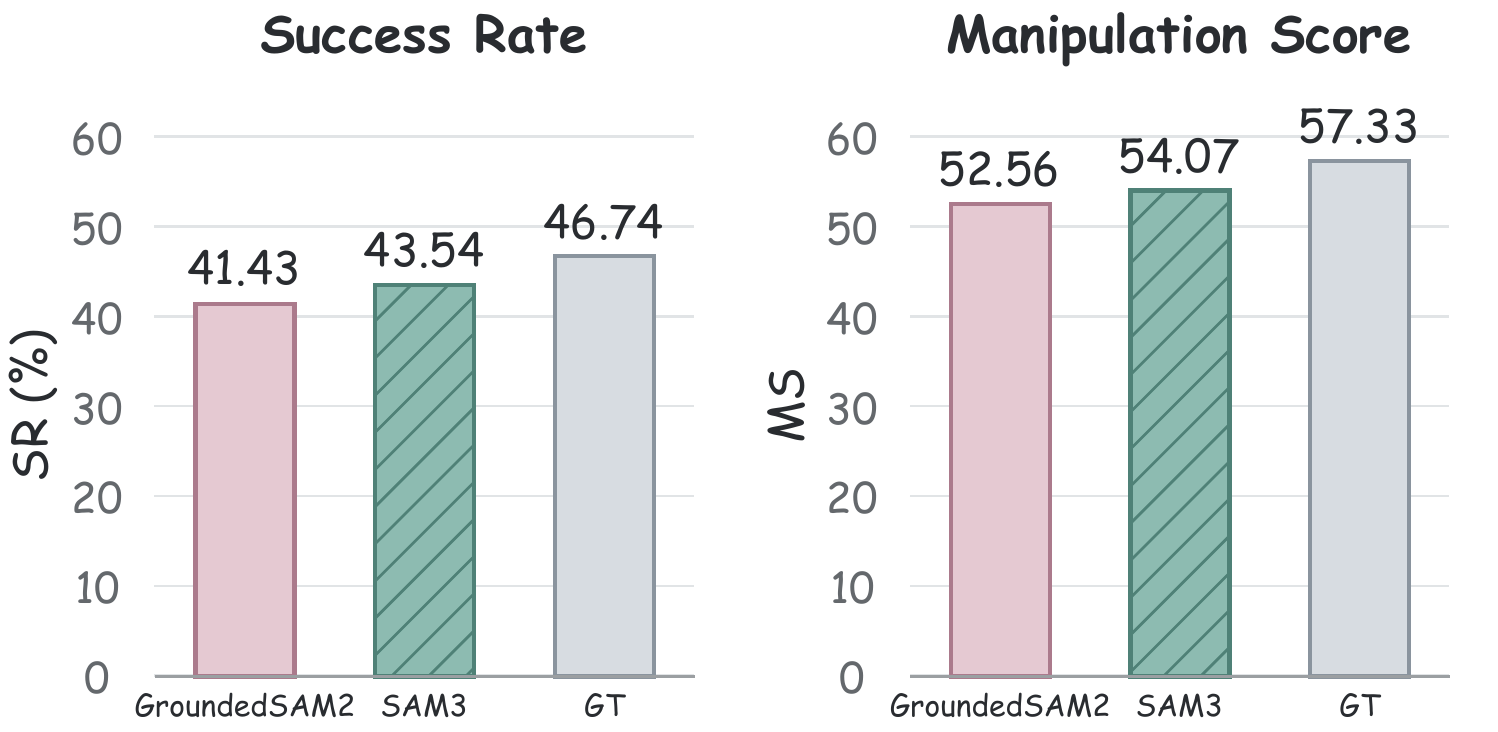}}
    \caption{Ablation of test-time mask sources on DOMINO. We evaluate success rate (left) and manipulation score (right) over 3,500 trials per source, with all model parameters fixed and only the observed masks replaced.}
    \label{fig:mask_sources}
    \vspace{-1.0em}
\end{figure}

\subsection{Ablation Studies}

We conduct ablation studies on DOMINO to examine the role of trajectory conditioning and the design choices for one-step deployment, reporting SR and MS under the settings specified below.

\subsubsection{Robustness to Test-Time Mask Sources}
We evaluate the sensitivity of TrajWAM to the source of target-object masks at inference. Training uses the precomputed ground-truth masks provided with the dataset. At test time, we compare three sources for the observed mask history: simulator-provided ground-truth masks, SAM3~\cite{carion2026sam3segmentconcepts}, and GroundedSAM2~\cite{liu2024groundingdinomarryingdino,ravi2024sam2segmentimages}. The trained trajectory predictor and downstream policy remain unchanged across these evaluations. For all three sources, future mask trajectories are predicted from the observed history.
As shown in Fig.~\ref{fig:mask_sources}, TrajWAM maintains competitive performance across the evaluated mask sources without source-specific retraining. 
Replacing SAM3 with Grounded SAM 2 further decreases SR by 2.11 percentage points and MS by 1.51 points. These results suggest that TrajWAM is relatively robust to the evaluated mask sources: despite being trained with ground-truth masks, it retains most of its manipulation performance when deployed with off-the-shelf segmentation models.

\subsubsection{Effect of Test-Time Trajectory Conditions}
We compare three trajectory conditions at inference on a fixed subset of 10 DOMINO tasks: \textit{Predicted} uses the predicted future masks; \textit{Static} repeats the current mask; and \textit{Black} uses all-zero masks. 
All other settings are kept unchanged.
As shown in Table~\ref{tab:ablation_summary}(a), \textit{Predicted} outperforms \textit{Static} and \textit{Black} by 26.40 and 15.30 percentage points in SR, respectively, supporting the benefit of predicted motion conditions for the deployed policy. \textit{Static} performs worse than \textit{Black}, possibly because repeating the current mask imposes an incorrect stationary trajectory that misguides future imagination as the target moves. \textit{Black}, although providing no explicit trajectory information, avoids this persistent misleading guidance.

\begin{table}[t]
    \centering
    \caption{Ablation studies on DOMINO.}
    \label{tab:ablation_summary}
    \footnotesize
    \setlength{\tabcolsep}{4pt}
    \begin{tabular*}{\columnwidth}{@{\extracolsep{\fill}}lrr@{}}
        \toprule
        Variant & SR (\%) & MS \\
        \midrule
        \multicolumn{3}{@{}l}{\textit{(a) Trajectory conditions (10 tasks)}} \\
        Static & 20.90 & 33.72 \\
        Black (all-zero) & 32.00 & 44.32 \\
        Predicted & \textbf{47.30} & \textbf{58.32} \\
        \midrule
        \multicolumn{3}{@{}l}{\textit{(b) One-step action distillation (12 tasks)}} \\
        Undistilled (1-step inference) & 7.39 & 15.82 \\
        DMD & 1.08 & 19.10 \\
        CD & 18.42 & 36.33 \\
        Action rollout (ours) & \textbf{42.25} & \textbf{53.38} \\
        \midrule
        \multicolumn{3}{@{}l}{\textit{(c) Trajectory adapter realignment (10 tasks)}} \\
        Without realignment & 42.20 & 52.76 \\
        With realignment & \textbf{43.20} & \textbf{54.11} \\
        \bottomrule
    \end{tabular*}
    \vspace{-1.0em}
\end{table}

\subsubsection{Comparison of Action Distillation Strategies}
We investigate how different action distillation strategies affect manipulation performance under one-step inference. All variants are evaluated on the same fixed subset of 12 DOMINO tasks, with all other settings kept unchanged. 
Observed mask histories are obtained from simulator-provided ground-truth masks, while future masks are still predicted by the mask predictor. 
We compare four variants: an undistilled action model, consistency distillation (CD)~\cite{song2023consistencymodels}, distribution matching distillation (DMD)~\cite{yin2024onestepdiffusiondistributionmatching}, and our action rollout distillation. The undistilled baseline uses the action teacher trained for 10-step sampling but directly performs inference with a single denoising step. CD and DMD train one-step action students using their respective distillation objectives.
As shown in Table~\ref{tab:ablation_summary}(b), action rollout distillation achieves the highest SR and MS, exceeding the strongest alternative CD by 23.83 percentage points and 17.05 points, respectively. The undistilled baseline achieves only 7.39\% SR, indicating that directly applying the multi-step-trained action model with a single denoising step is insufficient for effective manipulation in this setting. DMD achieves only 1.08\% SR; one possible explanation is that inaccuracies in its learned action scores compromise the distribution-matching updates, making it difficult to preserve the precise action timing and spatial alignment required for dynamic manipulation. 

\subsubsection{Effect of Trajectory Adapter Realignment}
We evaluate trajectory adapter realignment on a fixed subset of 10 DOMINO tasks. The variant without realignment directly pairs the trajectory adapter trained with the multi-step video model with the distilled one-step backbone. As shown in Table~\ref{tab:ablation_summary}(c), realignment improves SR from 42.20\% to 43.20\% and MS from 52.76 to 54.11, suggesting that adapting the Trajectory ControlNet to the distilled backbone helps compensate for changes in its intermediate representations, preserving effective trajectory guidance under one-step inference.
\section{Conclusion}
We presented TrajWAM, an efficient trajectory-guided world action model for dynamic robotic manipulation. Predicted object-mask trajectories guide future video generation, providing motion-aware representations for action prediction. A decoupled distillation framework enables one-step video and action generation while preserving this conditioning pathway. TrajWAM achieves a 43.54\% success rate on DOMINO with an inference latency of 108.5 ms and outperforms evaluated baselines on real-world dynamic tasks, demonstrating the effectiveness of trajectory-guided future imagination for efficient anticipatory control. However, the current experiments focus primarily on independently moving target objects; extending trajectory forecasting to action-dependent object dynamics remains an important direction toward more general dynamic manipulation.

\appendix
\renewcommand{\thesection}{\Alph{section}}
\renewcommand{\thesubsection}{\thesection.\arabic{subsection}}

\subsection{Experimental Setup and Evaluation Protocols}
\label{app:experimental_details}

This section specifies the task categories, ablation protocols, and real-world evaluation settings.

\subsubsection{DOMINO Task Categories}
\label{app:domino_task_categories}

Table~\ref{tab:appendix_task_categories} lists the 35 DOMINO tasks under the benchmark's official categories of Dynamic Interception and Dynamic Tracking~\cite{fang2026generalizableroboticmanipulationdynamic}. In Dynamic Interception tasks, the robot grasps a moving target before performing subsequent manipulation. In Dynamic Tracking tasks, continued adaptation to target motion is required throughout execution.

\subsubsection{Ablation Evaluation Protocols}
\label{app:ablation_protocols}

The ablations in Table~\ref{tab:ablation_summary} use simulator-provided masks for the observed history and the fixed task subsets listed in Table~\ref{tab:appendix_ablation_tasks}. The trajectory-condition and adapter-realignment experiments use different subsets of 10 tasks, with 100 evaluation episodes per task for each variant.

In the trajectory-condition ablation, \textit{Predicted} uses the predicted future masks, \textit{Static} repeats the current object mask, and \textit{Black} replaces the mask sequence with all-zero masks before VAE encoding. All three variants retain the trajectory adapter and use the same contact-based conditioning rule described below. Thus, \textit{Black} changes the adapter's mask input; it does not remove the adapter.

For the action-distillation comparison, we select 12 tasks from those with completed evaluations of our method. We stratify the candidate tasks using the success rates of a previously evaluated teacher policy: below $25\%$, $25$--$55\%$, and above $55\%$. Four tasks are selected from each group to cover different manipulation objectives. This subset is used only for evaluation; action distillation uses the full DOMINO demonstration mixture.

\subsubsection{Real-World Evaluation Protocol}
\label{app:real_world_protocol}

For real-world evaluation, the low, medium, and high speeds are 4, 7, and 10\,cm/s for the conveyor belt and $5^\circ$/s, $16^\circ$/s, and $24^\circ$/s for the rotating disc, respectively. Each task--speed combination includes 20 trials for each of four object categories, with randomized initial poses, giving 240 trials per task and 480 trials per method. A trial is successful if the robot grasps and lifts the target object.

\begin{table}[t]
    \centering
    \caption{Official task categories of DOMINO.}
    \label{tab:appendix_task_categories}
    \footnotesize
    \begin{tabular}{@{}ll@{}}
        \toprule
        Dynamic Interception (19) & Dynamic Tracking (16) \\
        \midrule
        Adjust Bottle & Beat Block Hammer \\
        Dump Bin Bigbin & Click Alarmclock \\
        Grab Roller & Click Bell \\
        Handover Block & Move Pillbottle Pad \\
        Handover Mic & Move Stapler Pad \\
        Hanging Mug & Place Bread Basket \\
        Move Can Pot & Place Container Plate \\
        Move Playing Card Away & Place Empty Cup \\
        Place A2B Left & Place Fan \\
        Place A2B Right & Place Mouse Pad \\
        Place Bread Skillet & Place Object Scale \\
        Place Can Basket & Place Object Stand \\
        Place Object Basket & Place Phone Stand \\
        Put Bottles Dustbin & Place Shoe \\
        Put Object Cabinet & Press Stapler \\
        Rotate QRcode & Stamp Seal \\
        Scan Object &  \\
        Shake Bottle &  \\
        Shake Bottle Horizontally &  \\
        \bottomrule
    \end{tabular}
\end{table}

\subsection{Model and Inference Implementation Details}
\label{app:model_inference_details}

The video and action branches follow the Wan2.2-derived architecture and are initialized from the pretrained Fast-WAM-Joint checkpoint, as specified in Sec.~\ref{Experimental Setup}.

\subsubsection{Object-Centric Trajectory Predictor}
\label{app:trajectory_prediction_details}

For a sample anchored at frame $t$, the predictor receives masks at $(t-12,t-8,t-4,t)$ and predicts masks at $(t+4,t+8,\ldots,t+32)$, where indices refer to frames in the recorded demonstration. Training windows advance by one frame and are retained only when the complete history and future lie within the same demonstration. Masks are resized to $120\times160$ (height $\times$ width) using nearest-neighbor interpolation and binarized at $0.5$.

We split the 1,750 DOMINO demonstration videos into 1,575 training videos and 175 validation videos using random seed 42. Splitting at the video level prevents windows from the same demonstration from appearing in both sets. We use AdamW with weight decay $10^{-4}$ and gradient norm clipping at 1.0, and select the checkpoint with the lowest validation loss. The training schedule and prediction loss are specified in Secs.~\ref{Experimental Setup} and~\ref{sec:trajectory_prediction}, respectively.

The four history masks are stacked along the channel dimension and processed by the U-Net in Table~\ref{tab:trajectory_predictor_architecture}. Each convolutional block contains two $3\times3$ convolutions, each followed by group normalization with eight groups and a SiLU activation. Downsampling uses stride-two $3\times3$ convolutions. Each decoder stage upsamples with a stride-two $2\times2$ transposed convolution, concatenates the corresponding encoder features, and applies a convolutional block. A final $1\times1$ convolution produces eight logit maps, one per future time step.

\begin{table*}[t]
    \centering
    \caption{Evaluation subsets for the ablations in Table~\ref{tab:ablation_summary}. Task names follow the DOMINO identifiers.}
    \label{tab:appendix_ablation_tasks}
    \footnotesize
    \begin{tabular}{@{}p{0.23\textwidth}p{0.73\textwidth}@{}}
        \toprule
        Ablation & Tasks \\
        \midrule
        (a) Trajectory conditions (10) & \texttt{click\_bell}, \texttt{press\_stapler}, \texttt{move\_pillbottle\_pad}, \texttt{place\_shoe}, \texttt{rotate\_qrcode}, \texttt{move\_can\_pot}, \texttt{grab\_roller}, \texttt{handover\_block}, \texttt{put\_bottles\_dustbin}, \texttt{place\_object\_basket} \\
        \midrule
        (b) Action distillation (12) & \texttt{click\_bell}, \texttt{press\_stapler}, \texttt{place\_object\_scale}, \texttt{place\_shoe}, \texttt{move\_pillbottle\_pad}, \texttt{place\_empty\_cup}, \texttt{rotate\_qrcode}, \texttt{move\_can\_pot}, \texttt{grab\_roller}, \texttt{handover\_block}, \texttt{put\_bottles\_dustbin}, \texttt{place\_object\_basket} \\
        \midrule
        (c) Adapter realignment (10) & \texttt{click\_alarmclock}, \texttt{move\_stapler\_pad}, \texttt{place\_mouse\_pad}, \texttt{place\_phone\_stand}, \texttt{dump\_bin\_bigbin}, \texttt{move\_pillbottle\_pad}, \texttt{rotate\_qrcode}, \texttt{place\_a2b\_left}, \texttt{grab\_roller}, \texttt{place\_object\_basket} \\
        \bottomrule
    \end{tabular}
\end{table*}

\begin{table}[t]
    \centering
    \caption{Trajectory predictor architecture.}
    \label{tab:trajectory_predictor_architecture}
    \footnotesize
    \setlength{\tabcolsep}{5pt}
    \begin{tabular}{@{}lcc@{}}
        \toprule
        Stage & Output channels & Spatial size \\
        \midrule
        History input & 4 & $120\times160$ \\
        Input block & 48 & $120\times160$ \\
        Encoder stage 1 & 96 & $60\times80$ \\
        Encoder stage 2 & 192 & $30\times40$ \\
        Bottleneck & 192 & $30\times40$ \\
        Decoder stage 1 & 96 & $60\times80$ \\
        Decoder stage 2 & 48 & $120\times160$ \\
        Output projection & 8 & $120\times160$ \\
        \bottomrule
    \end{tabular}
\end{table}

\subsubsection{Input and Conditioning Representations}
\label{app:input_conditioning}

Predicted probabilities are thresholded at $0.5$, and the latest history mask is prepended to the eight predictions to form a nine-frame conditioning sequence. For the DOMINO camera layout, each mask is resized to the $256\times320$ third-person region and padded with zeros over the $128\times320$ wrist-view region. The resulting $384\times320$ frames are replicated across three channels and mapped from $[0,1]$ to $[-1,1]$ before video-VAE encoding.

For DOMINO, the proprioceptive input consists of the current 12 arm-joint positions and two gripper states. After z-score normalization using the saved dataset statistics, a linear proprioceptive encoder projects this 14-dimensional vector into a single 4096-dimensional token, which is appended to the language-token sequence to form the context supplied to both the video and action branches.

\subsubsection{Online Mask History}
\label{app:online_mask_history}

During DOMINO evaluation, observed masks are stored with their observation-step indices in a buffer that is cleared at the start of each episode. For each requested history index, we retrieve the most recent mask at or before that index; indices preceding the first observation are filled with the earliest available mask. If segmentation returns no mask, we reuse the previous mask, or an all-zero mask if the buffer is empty. Only observed masks are added to this buffer.

\subsubsection{Contact-Based Trajectory Conditioning}
\label{app:contact_conditioning}

In DOMINO Dynamic Interception tasks, the target's motion is largely governed by the robot after grasping, making extrapolation of its preceding free motion less relevant. We therefore use simulator-reported contact between the gripper and the target as a proxy for the transition from interception to subsequent manipulation. Contact is checked before each policy update; upon the first detection, the trajectory-adapter residuals are set to zero for the remainder of the episode. In Dynamic Tracking tasks, trajectory conditioning remains active throughout execution. The same rule is applied to all three variants in the trajectory-condition ablation.

\subsection{Distillation Details}
\label{app:distillation_deployment}

The following settings supplement the distillation objectives in Sec.~\ref{sec:decoupled_one_step_distillation} and the training schedules in Sec.~\ref{Experimental Setup}. We first describe the three stages of our distillation pipeline, followed by the action-distillation baselines.

Video backbone distillation and trajectory adapter realignment share the following optimization settings: AdamW with $(\beta_1,\beta_2)=(0,0.999)$, weight decay $0.01$, gradient norm clipping at $10$, and BF16 precision. Learning rates are constant, without warm-up. For these two stages, training iterations refer to the outer loop, with generator or adapter updates performed once every five iterations.

\subsubsection{Video Backbone Distillation}
\label{app:video_distillation_realignment}

For video DMD, the generator, frozen real-score model, and trainable fake-score model are initialized from the same pretrained video weights. Generation uses a single Euler update from normalized noise level $1$ to $0$. For score evaluation, noise levels are sampled uniformly from $[0.02,0.98]$ in both generator and fake-score updates. The observed-frame latent remains fixed, and losses are computed only over valid future positions. Generator and fake-score updates use separately drawn batches.

We initialize an exponential moving average (EMA) of the generator at training iteration 200, with decay $0.99$, and update it after each subsequent generator optimization step. The averaged weights are used for trajectory adapter realignment and deployment.

\subsubsection{Trajectory Adapter Realignment}
\label{app:adapter_realignment}

Realignment uses demonstration windows in which trajectory conditioning is active, with conditions constructed from ground-truth object masks. The coefficient of $\mathcal{L}_{\mathrm{rec}}$ increases linearly from $0.1$ to $0.35$ over 1,000 training iterations; the coefficients of $\mathcal{L}_{\mathrm{adv}}$ and $\mathcal{L}_D$ are both $1$. Reconstruction is evaluated only over valid future latent positions. Adapter and discriminator updates use separately drawn batches at a ratio of $1:5$, with a learning rate of $2\times10^{-6}$ for both.

The discriminator reads features after the 14th, 22nd, and 30th transformer blocks of its frozen video backbone. At each depth, a learned query aggregates language and observed-image features, attends successively to trajectory and video tokens, and passes through a feed-forward block. Attention uses 24 heads and feature width 3,072; the feed-forward hidden width is 14,336. The three query outputs are concatenated into a 9,216-dimensional vector and passed through layer normalization and a two-layer MLP with hidden width 3,072 and a SiLU activation to produce a scalar logit.

For adversarial evaluation in both adapter and discriminator updates, we sample $u\sim\mathcal{U}(0,1)$, transform it as $\tilde{\tau}=5u/(1+4u)$, and clamp the result to $[0.02,0.98]$. During discriminator updates, ground-truth and generated future latents are perturbed using the same noise realization and noise level, while the observed-frame latent remains unperturbed.

\subsubsection{Action Rollout Distillation}
\label{app:action_rollout_distillation}

After video backbone distillation and trajectory adapter realignment, the one-step video generator and trajectory adapter are fixed while the student action branch is optimized. The teacher uses 10 video and 10 action denoising steps, whereas the student uses one step for each modality. Teacher and student receive the same observation, language instruction, trajectory condition, and initial video and action noises. Each model produces its own video representation through its respective rollout. Following the training schedule in Sec.~\ref{Experimental Setup}, we train the student action branch for 2,000 steps with a global batch size of 4 and a constant learning rate of $10^{-6}$.

The student is trained to match the final action chunk produced by the teacher's multi-step rollout, using the action rollout objective in Sec.~\ref{sec:decoupled_one_step_distillation}. Padded action positions are excluded from the loss. This objective supervises the final action prediction under the student's one-step video representation. Distillation uses the full DOMINO demonstration mixture; the 12-task subset in Table~\ref{tab:appendix_ablation_tasks} is used only for evaluation.

\subsubsection{Action Distillation Baselines}
\label{app:action_distillation}

For the comparison in Table~\ref{tab:ablation_summary}(b), CD and DMD initialize their students from the same trajectory-conditioned action teacher and optimize the full action branch while freezing the video generator, trajectory adapter, and proprioceptive encoder. The two baselines share the same one-step video checkpoint, and all three distilled variants use the same trajectory-adapter checkpoint. Table~\ref{tab:action_distillation_config} lists the CD and DMD optimization settings.

\begin{table}[t]
    \centering
    \caption{Training settings for the action CD and DMD baselines.}
    \label{tab:action_distillation_config}
    \footnotesize
    \begin{tabular}{@{}ll@{}}
        \toprule
        Setting & Value \\
        \midrule
        Training steps & 5,000 \\
        Global batch size & 1 \\
        Optimizer & AdamW \\
        Learning rate & $5\times10^{-6}$ \\
        Learning-rate schedule & Linear warm-up, then constant \\
        Warm-up steps & 100 \\
        Adam coefficients $(\beta_1,\beta_2)$ & $(0.9,0.95)$ \\
        Weight decay & 0 \\
        Gradient norm clipping & 2.0 \\
        EMA decay & 0.995 \\
        Training precision & BF16 \\
        \bottomrule
    \end{tabular}
\end{table}

Within each CD or DMD training sample, the student, frozen teacher, and target or auxiliary network share the future video latents generated by the fixed one-step video model. Padded action positions are excluded from the losses, and neither baseline adds an auxiliary ground-truth action regression loss.

For CD~\cite{song2023consistencymodels}, a demonstration action chunk is perturbed at normalized noise level $\sigma\sim\mathcal{U}(0,1)$, and the frozen teacher performs one Euler update to $\sigma'=\max(\sigma-0.5,0)$. The student estimates the clean action chunk from the initial noisy input, while an EMA target network estimates it from the teacher-updated input. Their discrepancy is penalized by $\rho(e)=\sqrt{e^2+c^2}-c$, with $c=10^{-3}$, averaged over valid action entries. The target network is initialized from the student and updated by EMA after each optimization step. The teacher update constructs a training target; inference uses a single action denoising step.

\begin{table}[t]
    \centering
    \caption{Per-task results of TrajWAM on DOMINO.}
    \label{tab:domino_sam3_all35_per_task}
    \footnotesize
    \setlength{\tabcolsep}{4pt}
    \begin{tabular}{@{}lrr@{}}
        \toprule
        Task & SR (\%) & MS \\
        \midrule
        \texttt{adjust\_bottle} & 51.00 & 67.17 \\
        \texttt{beat\_block\_hammer} & 46.00 & 55.83 \\
        \texttt{click\_alarmclock} & 14.00 & 19.24 \\
        \texttt{click\_bell} & 4.00 & 17.28 \\
        \texttt{dump\_bin\_bigbin} & 33.00 & 50.41 \\
        \texttt{grab\_roller} & 52.00 & 68.84 \\
        \texttt{handover\_block} & 67.00 & 85.50 \\
        \texttt{handover\_mic} & 56.00 & 64.86 \\
        \texttt{hanging\_mug} & 40.00 & 54.06 \\
        \texttt{move\_can\_pot} & 71.00 & 81.49 \\
        \texttt{move\_pillbottle\_pad} & 35.00 & 41.00 \\
        \texttt{move\_playingcard\_away} & 40.00 & 49.71 \\
        \texttt{move\_stapler\_pad} & 18.00 & 29.95 \\
        \texttt{place\_a2b\_left} & 49.00 & 59.56 \\
        \texttt{place\_a2b\_right} & 55.00 & 68.35 \\
        \texttt{place\_bread\_basket} & 28.00 & 36.28 \\
        \texttt{place\_bread\_skillet} & 48.00 & 55.46 \\
        \texttt{place\_can\_basket} & 78.00 & 88.89 \\
        \texttt{place\_container\_plate} & 76.00 & 78.10 \\
        \texttt{place\_empty\_cup} & 41.00 & 48.19 \\
        \texttt{place\_fan} & 9.00 & 15.06 \\
        \texttt{place\_mouse\_pad} & 26.00 & 38.21 \\
        \texttt{place\_object\_basket} & 74.00 & 82.23 \\
        \texttt{place\_object\_scale} & 12.00 & 22.40 \\
        \texttt{place\_object\_stand} & 21.00 & 28.59 \\
        \texttt{place\_phone\_stand} & 26.00 & 40.71 \\
        \texttt{place\_shoe} & 35.00 & 43.01 \\
        \texttt{press\_stapler} & 3.00 & 15.66 \\
        \texttt{put\_bottles\_dustbin} & 69.00 & 76.66 \\
        \texttt{put\_object\_cabinet} & 49.00 & 67.58 \\
        \texttt{rotate\_qrcode} & 44.00 & 52.44 \\
        \texttt{scan\_object} & 24.00 & 48.91 \\
        \texttt{shake\_bottle} & 89.00 & 91.16 \\
        \texttt{shake\_bottle\_horizontally} & 91.00 & 92.98 \\
        \texttt{stamp\_seal} & 50.00 & 56.75 \\
        \midrule
        \textbf{Overall} & \textbf{43.54} & \textbf{54.07} \\
        \bottomrule
    \end{tabular}
\end{table}

For the action DMD baseline~\cite{yin2024onestepdiffusiondistributionmatching}, the student generates an action chunk from Gaussian noise in one step. The generated chunk is perturbed at a noise level sampled uniformly from $[0.02,0.98]$, and the frozen teacher and an auxiliary action network produce clean-action estimates. The auxiliary-minus-teacher difference is divided by the mean absolute difference between the generated chunk and the teacher estimate, with the denominator bounded below by $10^{-6}$. Subtracting this normalized difference from the generated chunk defines a stop-gradient regression target, against which the student is trained with squared error.

For the action DMD baseline in Table~\ref{tab:ablation_summary}(b), the auxiliary action network is initialized from pretrained action weights and updated by an exponential moving average of the student parameters with decay $0.995$. It has no separate optimizer and is not trained with a denoising objective on student-generated action samples. This baseline therefore uses a DMD-style distribution-matching update with an EMA auxiliary network, which differs from the independently trained fake-score model used in standard DMD and in our video distillation stage. The reported action DMD results characterize this implementation.

\subsection{Additional Results and Qualitative Analysis}
\label{app:additional_results}

\subsubsection{Per-Task Results on DOMINO}
\label{app:per_task_results}

Table~\ref{tab:domino_sam3_all35_per_task} expands the aggregate DOMINO results in Table~\ref{table_domino}. TrajWAM uses SAM3 for the observed mask history and is evaluated on 100 episodes per task, totaling 3,500 episodes. Averaged across tasks, SR is $43.54\%$ and MS is $54.07$.

\subsubsection{Trajectory Prediction Visualization}
\label{app:trajectory_visualization}

\begin{figure*}[t]
    \centering
    \includegraphics[width=\textwidth]{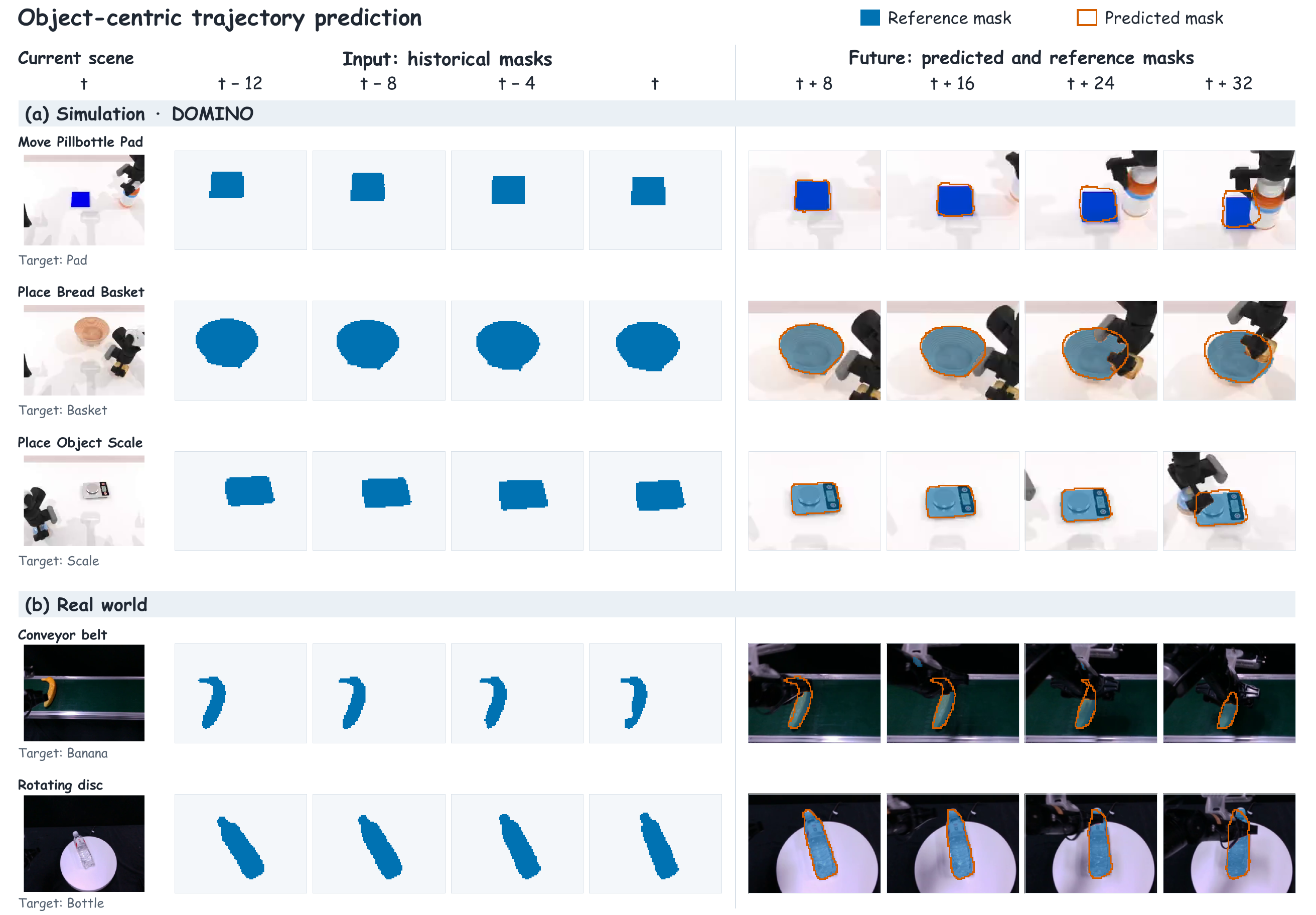}
    \caption{Object-centric trajectory prediction on three DOMINO tasks and two real-world grasping sequences. Each row shows the current scene, four input masks, and four of the eight future predictions. Blue regions denote reference masks; orange contours denote predictions. Future overlays use the corresponding recorded RGB frames. References are dataset ground-truth masks for DOMINO and SAM3 segmentations for the real-world sequences. Time labels indicate offsets in source-video frames relative to $t$. All temporal panels within a row share a fixed crop; the current scene retains the full field of view.}
    \label{fig:appendix_mask_prediction}
\end{figure*}

Figure~\ref{fig:appendix_mask_prediction} shows predictions from four observed masks in DOMINO training demonstrations and real-world sequences. The DOMINO targets are the receiving pad, basket, and scale, illustrating that the predicted trajectory can describe a moving placement destination. The real-world targets are the objects to be grasped. Predictions broadly follow target displacement and changes in orientation, with visible position and contour discrepancies in some frames.

\subsubsection{Video Prediction Visualization}
\label{app:video_visualization}

\begin{figure*}[t]
    \centering
    \includegraphics[width=\textwidth,height=0.8\textheight,keepaspectratio]{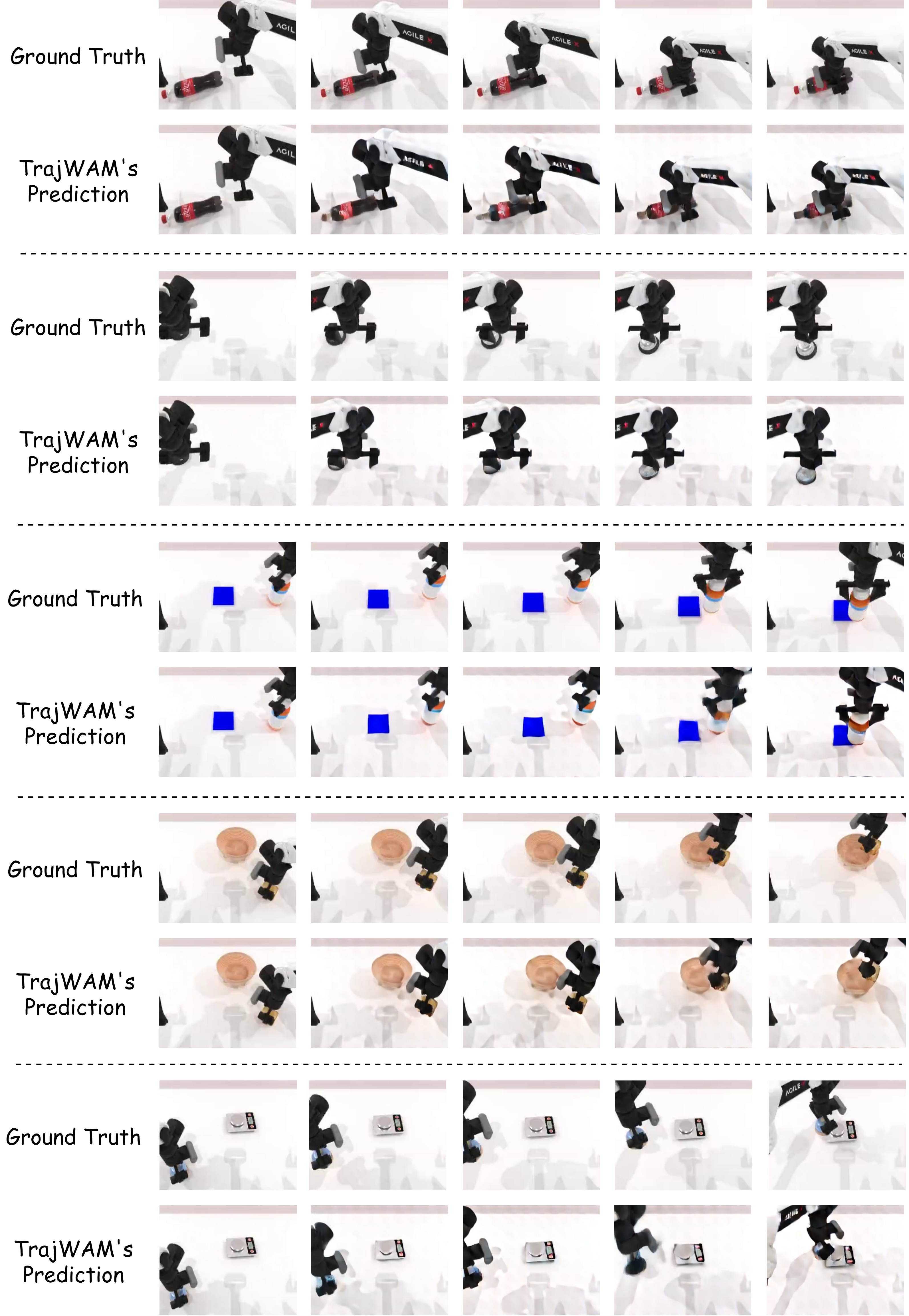}
    \caption{Video predictions on DOMINO training demonstrations. From top to bottom: Adjust Bottle, Click Bell, Move Pillbottle Pad, Place Bread Basket, and Place Object Scale. Each pair of rows shows the recorded sequence above the prediction, with time progressing from left to right. The leftmost column is the conditioning frame. The video generator uses one denoising step and ground-truth future object masks.}
    \label{fig:appendix_video_sim}
\end{figure*}

\begin{figure*}[t]
    \centering
    \includegraphics[width=\textwidth,height=0.8\textheight,keepaspectratio]{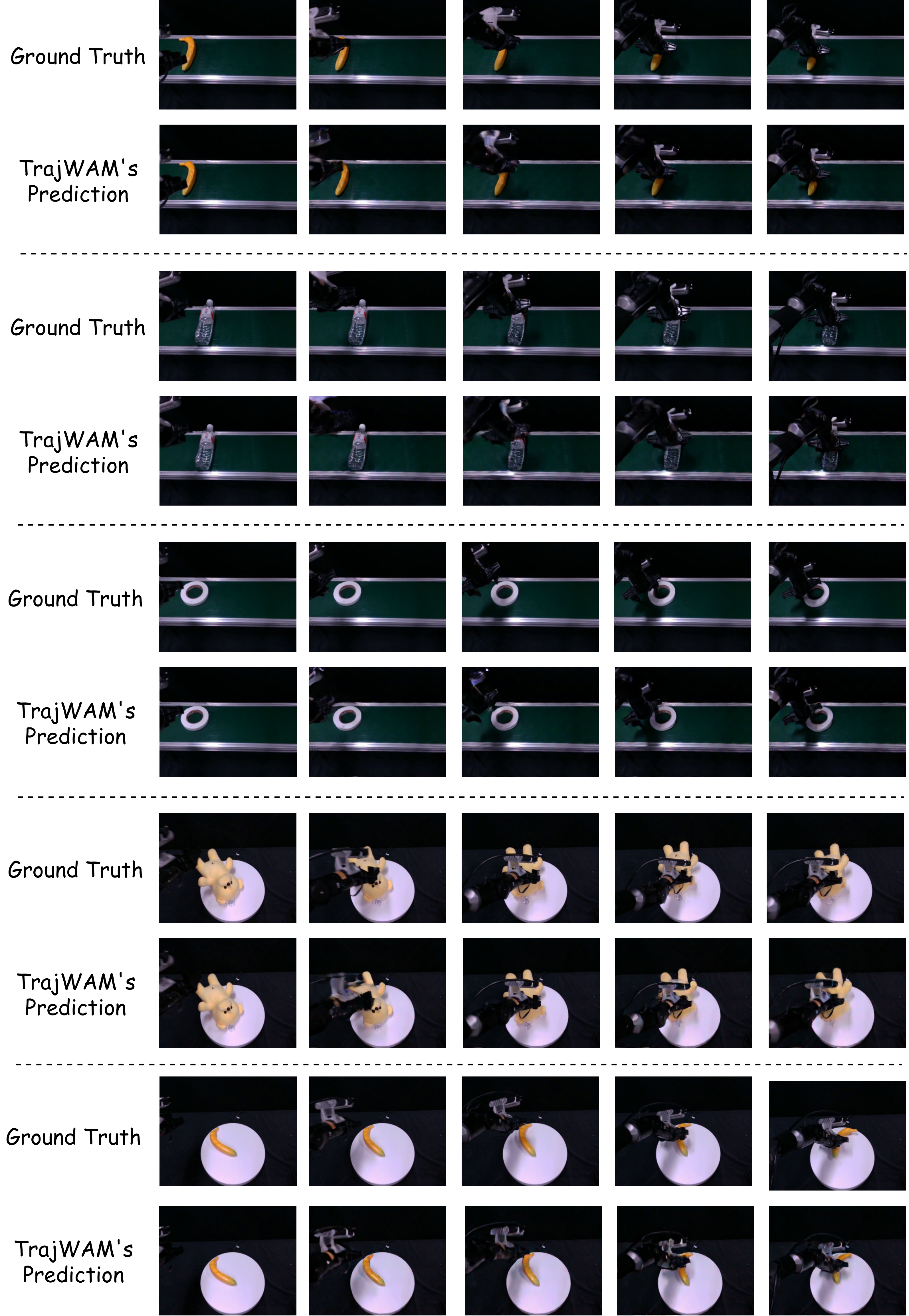}
    \caption{Video predictions for real-world dynamic grasping. The first three examples show conveyor-belt grasping with a banana, a bottle, and a roll of tape; the last two show rotating-disc grasping with a plush toy and a banana. Each pair of rows compares the recorded sequence with the prediction at corresponding time points, ordered from left to right.}
    \label{fig:appendix_video_real}
\end{figure*}

Figures~\ref{fig:appendix_video_sim} and~\ref{fig:appendix_video_real} compare predicted videos with recorded sequences. The DOMINO examples use ground-truth future object masks from training demonstrations to condition the distilled one-step generator, illustrating video generation under supplied trajectory conditions. The real-world examples cover conveyor-belt and rotating-disc grasping with objects of different shapes and appearances. In the displayed sequences, predictions capture target motion and the gripper's approach, with differences in object appearance and spatial alignment relative to the recordings.

\end{document}